\documentclass[fleqn,10pt]{wlpeerj}

\usepackage{algorithmic, algorithm, hyperref, multirow, booktabs, tabularray, array, graphicx}
\usepackage{url,hyperref,lineno,microtype}
\usepackage[onehalfspacing]{setspace}

\title{Label-efficient multi-organ segmentation with a diffusion model}

\author[1,2,\#]{Yongzhi Huang}
\author[3,\#]{Fengjun Xi}
\author[2]{Liyun Tu}
\author[1]{Jinxin Zhu}
\author[1]{Haseeb Hassan}
\author[1]{Liyilei Su}
\author[4]{Yun Peng}
\author[1]{Jingyu Li}
\author[3,*]{Jun Ma}
\author[1,*]{Bingding Huang}

\affil[1]{College of Big Data and Internet, Shenzhen Technology University, Shenzhen, 518118, China}
\affil[2]{School of Artificial Intelligence, Beijing University of Posts and Telecommunications, Beijing, 100876, China}
\affil[3]{Department of Radiology, Beijing Tiantan Hospital, Capital Medical University, Beijing, China}
\affil[4]{Department of Radiology, MOE Key Laboratory of Major Diseases in Children, Beijing Children’s Hospital, Capital Medical University, National Center for Children’s Health, Beijing 100045, China}
\corrauthor{Bingding Huang and Jun Ma}{huangbingding@sztu.edu.cn, majun@bjtth.org}

\begin{abstract}
Accurate segmentation of multiple organs in Computed Tomography (CT) images plays a vital role in computer-aided diagnosis systems. While various supervised learning approaches have been proposed recently, these methods heavily depend on a large amount of high-quality labeled data, which are expensive to obtain in practice. To address this challenge, we propose a label-efficient framework using knowledge transfer from a pre-trained diffusion model for CT multi-organ segmentation. Specifically, we first pre-train a denoising diffusion model on 207,029 unlabeled 2D CT slices to capture anatomical patterns. Then, the model backbone is transferred to the downstream multi-organ segmentation task, followed by fine-tuning with few labeled data. In fine-tuning, two fine-tuning strategies, linear classification and fine-tuning decoder, are employed to enhance segmentation performance while preserving learned representations. Quantitative results show that the pre-trained diffusion model is capable of generating diverse and realistic 256x256 CT images (Fréchet inception distance (FID): 11.32, spatial Fréchet inception distance (sFID): 46.93, F1-score: 73.1\%). Compared to state-of-the-art methods for multi-organ segmentation, our method achieves competitive performance on the FLARE 2022 dataset, particularly in limited labeled data scenarios. After fine-tuning with 1\% and 10\% labeled data, our method achieves dice similarity coefficients (DSCs) of 71.56\% and 78.51\%, respectively. Remarkably, the method achieves a DSC score of 51.81\% using only four labeled CT slices. These results demonstrate the efficacy of our approach in overcoming the limitations of supervised learning approaches that is highly dependent on large-scale labeled data.

\end{abstract}

\begin{document}

\flushbottom
\maketitle
\thispagestyle{empty}

\section*{Introduction}

Medical image segmentation is a critical task in medical imaging, as it enables accurate diagnosis and treatment planning for various diseases \cite{zhang2024challenges}.  Recent advancements in deep learning has gained popularity in medical image segmentation due to its ability to automatically learn relevant features from data and its superior performance compared to traditional segmentation  \cite{azad2024medical}. These approaches have demonstrated promising results in accurately segmenting organs and tissues from medical images, including Magnetic Resonance Imaging (MRI) and Computed Tomography (CT) scans \cite{ronneberger2015u, isensee2021nnu}.

However, accurate annotations for medical image segmentation, especially in real clinical scenarios, remains a challenging and time-consuming task requiring professional expertise \cite{karimi2020deep}. Furthermore, the cross-domain problem caused by different scanners and imaging protocols \cite{yanzhen2024exploring}, as well as the inevitable noise label introduced by experts, significantly decreases the quality of training data. These issues pose challenges to training robust and comprehensive models, which render supervised learning approaches impractical and highlight the necessity for methods that can learn effectively from limited labeled data. Semi-supervised and self-supervised learning techniques provide a promising training paradigm, utilizing limited labeled data and unlabeled data, or pre-training on unlabeled data and fine-tuning with labeled data. In these paradigms, generative models contribute significantly by generating high-quality synthetic data or learning robust feature representations, which has the potential to promote label-efficient learning in medical imaging.
 
The Denoising Diffusion Probabilistic Model (DDPM) is an advanced generative model known for its proficiency in generative tasks \cite{baranchuk2021label}. Existing works explored DDPM for generative pre-training, with learned feature representations transferred to various downstream tasks \cite{chung2023fast, la2022anatomically, kim2022diffusionclip}. However, it is less explored in medical imaging remains limited due to two main challenges: (1) the requirement of pre-training DDPM with medical image data from scratch, and (2) the lack of effective transfer and fine-tuning strategies for DDPM. To fill this gap, we propose a novel framework that pre-trains a segmentation network following the denoising process in DDPM, and then transfers the pre-trained network to downstream CT multi-organ tasks. Our main contributions are summarized as follows:

(1) We adapt the advanced generative model, DDPM, to be applicable to CT images. We pre-train a DDPM from scratch with 207,029 unlabeled 2D CT slices. To improve authenticity, we proposed a novel data preprocessing pipeline for synthesizing abdomen CT images. As a result, the pre-trained DDPM is capable of generating diverse and realistic 256x256 CT images, achieving the performance in terms of Fréchet inception distance (FID), spatial Fréchet inception distance (sFID), and F1-score, with values of 11.32, 46.93, and 73.1\%, respectively.

(2) We propose lightweight, end-to-end fine-tuning strategies to transfer the pre-trained network in DDPM to multi-organ segmentation tasks. We thoroughly evaluated the segmentation performance of our proposed method on the MICCAI FLARE2022 dataset. By leveraging generative pre-training on unlabeled CT images, the model achieves performance comparable to SOTA methods in multi-organ segmentation tasks after lightweight fine-tuning.

(3) Our proposed approach achieves superior performance in limited labeled data scenarios. After fine-tuning with 10\% and 1\% labeled data, our method achieves dice similarity coefficients (DSCs) of 78.51\% and 71.56\%, surpassing SOTA methods by more than 10\% and 5\%, respectively. Remarkably, the method achieves a DSC score of 51.81\% using only four labeled CT slices. 

\section*{Related work}

\subsection*{Pre-training Generative Models for Semantic Segmentation}

Generative models play a crucial role in unsupervised learning by capturing complex patterns in raw data without relying on labels. Over the past few years, Generative Adversarial Networks (GANs) \cite{goodfellow2020generative} have emerged as the dominant approach to generate images \cite{gui2021review}. Recently, diffusion models \cite{sohl2015deep, song2019generative, ho2020denoising, dhariwal2021diffusion} have emerged as a promising generation of generative models.

Ever since the introduction of generative models like GANs, researchers attempted to address semantic segmentation tasks by using these generative pre-trained models. Existing GAN-based segmentation methods \cite{bielski2019emergence, tritrong2021repurposing, li2021semantic} focus on improving segmentation performance by generating labeled image maps, extracting pixel-level representations, and capturing the joint distribution of image and label information, demonstrating the latent features of pre-trained GANs have potential advantages in semantic segmentation tasks. Obviously, it is easy to pose a question: can the DDPM, one of the emerging generative models, be applied to improve performance in semantic segmentation tasks like GANs? Quite a lot of representative works have verified this idea, including \cite{baranchuk2021label, brempong2022denoising, graikos2022diffusion} in natural images and \cite{wu2022medsegdiff, pinaya2022brain, guo2022accelerating} in medical images. 

Current DDPM-based methods for semantic segmentation can be categorized into three paradigms. The first line typically employs traditional classifiers (e.g., SVM or MLPs) to project the feature representations extracted from pre-trained generative models to segmentation masks \cite{tritrong2021repurposing, li2021semantic, baranchuk2021label}. However, these methods require manual hyperparameter selection for feature extraction in specific downstream tasks. The second follows the conditional generation pipeline by DDPM, integrating segmentation annotations as priors into the vanilla DDPM iterative sampling process \cite{graikos2022diffusion, wu2022medsegdiff, guo2022accelerating}. The limitation of these methods is the inference speed, where one needs to traverse all diffusion steps to output the mask. The third line, based on self-supervised learning or semi-supervised learning, utilizes generative pre-training via a DDPM process followed by fine-tuning \cite{brempong2022denoising}, or generates a large amount of synthetic medical images to train models with semi-supervised training approaches. The main challenge is to apply these methods to medical imaging, lying the need to pre-train DDPMs from scratch using medical image data.

\subsection*{Multi-organ Segmentation in Medical Imaging}

Since fully convolutional networks (FCNs) \cite{long2015fully} were initially introduced for semantic segmentation, deep learning-based methods for dense predictions have been explored for more than a decade. Advances in network architectures and training paradigms in computer vision have driven the rapid development of medical image segmentation methods. The most popular network architecture for medical image segmentation is U-Net \cite{ronneberger2015u}, which adopts an encoder-decoder architecture with skip connections to progressively down-sample and up-sample feature maps. Unlike natural image recognition tasks, organ segmentation methods in medical imaging require task-specific modifications to address domain-specific anatomical constraints or heterogeneous tissue characteristics. For example, \cite{liu2024seu2} proposed a novel squeeze-and-excitation attention mechanism that effectively captures multi-scale contextual features in CT images, particularly enhancing lesion boundary delineation in cases with ambiguous pathological margins. \cite{xu2024mcv} leveraged a hybrid Vision Transformers to model long-range, high-dimensional spatial dependencies in ultrasound images, demonstrating superior performance in segmenting intricate neural structures within noisy ultrasound imaging.

\subsection*{Label-efficient Segmentation}

Deep learning has transformed image segmentation by reducing reliance on labor-intensive pixel-level annotations. Label-efficient segmentation aims to achieve accurate dense predictions with minimal supervision \cite{shen2023survey}, where DDPM-based methods for semantic segmentation demonstrate unique potential through the two complementary branches: self-supervised learning and semi-supervised learning.

First, the DDPM training process inherently serves as a generative pre-training paradigm. Similar to contrastive learning in DINO \cite{caron2021emerging} or masked reconstruction in MAE \cite{he2022masked}, the noise prediction proxy task guides the model to capture semantic segmentation patterns in pre-training.  \cite{baranchuk2021label} reveals that intermediate denoising features encode segmentation-aware representations, enabling effective knowledge transfer from image-level pre-training to pixel-wise prediction tasks.
 
Second, by leveraging the generative capability of DDPM, synthetic medical images are generated to augment limited labeled datasets. Then, established semi-supervised learning frameworks or techniques, such as teacher-student architecture \cite{tarvainen2017mean}, self-training \cite{xie2022unsupervised} and pseudo-label refinement \cite{chen2021semi}, can be seamlessly incorporated into downstream segmentation tasks.

\section*{Methods}

Inspired by the recent success of DDPMs in image generation and its applications in semantic segmentation, we propose a label-efficient multi-organ framework for multi-organ segmentation by leveraging a diffusion model. By revisiting denoising objectives of generative pre-training in DDPM, we consider the DDPM process as pre-training, and transfer it to semantic segmentation tasks. 

Compared to other label-efficient approaches, our DDPM-based method has several advantages. First, we pre-train a DDPM from scratch on CT images rather than using a DDPM pre-trained on natural images. This fills in the domain gap between natural images and CT images, enabling the pre-trained model to segment multiple organs even fine-tuned with extremely limited labeled data. Second, we eliminate the need for task-specific hyperparameter configuration (e.g., block depth and diffusion step selection) required in \cite{baranchuk2021label}, as our approach directly leverages learned representations by a manner of end-to-end fine-tuning. Finally, we address the slow inference issue in iterative sampling pipelines by decoupling the generative process from mask prediction, enabling faster segmentation while preserving the DDPM inference framework. 

\begin{figure}[h!]
    \centering
    \includegraphics[width=\textwidth]{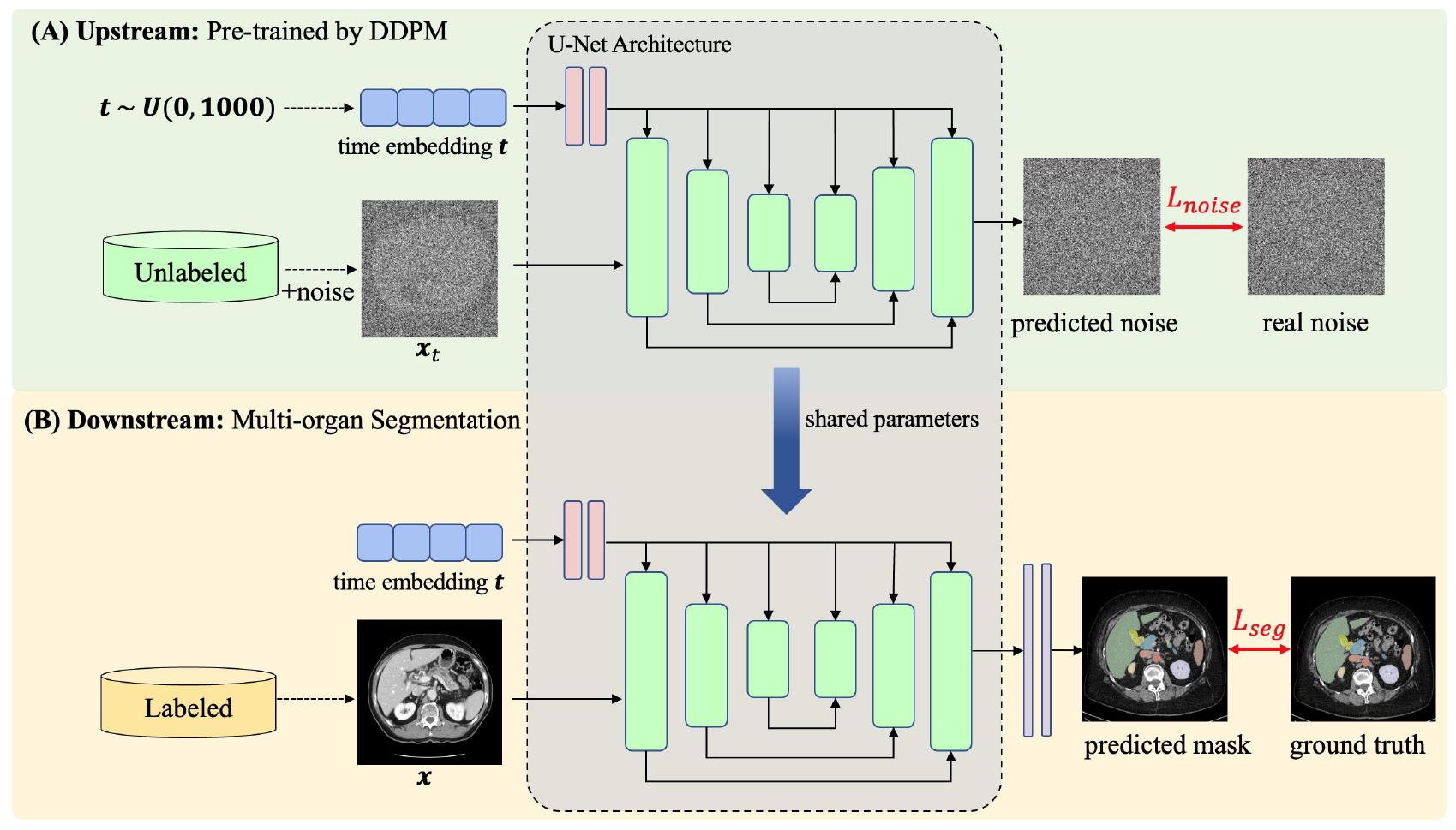}
    \caption{Our proposed method for CT multi-organ segmentation. It comprises the upstream DDPM pre-training task \textbf{(A)} and the downstream multi-organ segmentation task \textbf{(B)}.}
    \label{fig1: framework}
\end{figure}

As shown in Figure \ref{fig1: framework}, our proposed approach comprises two stages: \textbf{(A)} the upstream stage that pre-trains a model by DDPM, and \textbf{(B)} the downstream stage that transfers the pre-trained model to multi-organ segmentation tasks. In the upstream stage, the model is pre-trained to predict noise in noisy images using unlabeled CT slices. The U-Net network takes a noisy input image $\mathbf{x}_{t}$ and a time embedding \(t\), then learns to predict the underlying noise through a noise-prediction loss \(\mathcal{L}_{\text{noise}}\). In the downstream stage, the same U-Net architecture (with shared parameters) is applied and fine-tuned by annotations for multi-organ segmentation. Here, the network receives the original input image $\mathbf{x}$ and a fixed time embedding \(t\), producing a segmentation mask that is evaluated against ground truth masks using a segmentation loss \(\mathcal{L}_{\text{seg}}\).

\subsection*{Pre-training Models with DDPM}

\subsubsection*{Diffusion Model\label{section: diffusion model}}

DDPM consists of two primary processes: the sampling process and the diffusion process. In the sampling process, images are iteratively transformed from an initial state to a target state by applying Gaussian noise and continuous perturbations, generating diverse and high-quality images. On the other hand, the diffusion process is a Markov chain that gradually introduces noise to the original data, moving in the opposite direction of the sampling process until the signal becomes corrupted. DDPM can learn a data distribution, denoted as $p_{\theta}\left(\mathbf{x}_{0}\right)$, that effectively approximates a given data distribution, represented as $q_{}\left(\mathbf{x}_{t}\right)$. An advantage of DDPM is that progressive sampling is more efficient in generating data samples.

The forward process of DDPM involves gradually increasing the noise in the data:
\begin{equation}
\label{formulation: forward process}
q\left(\mathbf{x}_{t} \mid \mathbf{x}_{t-1}\right):=\mathcal{N}\left(\mathbf{x}_{t} ; \sqrt{1-\beta_{t}} \mathbf{x}_{t-1}, \beta_{t} \mathbf{I}\right)
\end{equation}
for some fixed variance schedule $\beta_{1}, \ldots, \beta_{t}$. Furthermore, noise samples $\mathbf{x}_{t}$ can be obtained directly from data $\mathbf{x}_{0}$:
\begin{equation}
q\left(\mathbf{x}_{t} \mid \mathbf{x}_{0}\right):=\mathcal{N}\left(\mathbf{x}_{t} ; \sqrt{\bar{\alpha}_{t}} \mathbf{x}_{0},\left(1-\bar{\alpha}_{t}\right) \mathbf{I}\right)
\end{equation}
where $\alpha_{t}:=1-\beta_{t}, \bar{\alpha}_{t}:=\prod_{s=1}^{t} \alpha_{s}$.

DDPMs transform noise $\mathbf{x}_{T} \sim \mathcal{N}(\mathbf{0}, \mathbf{I})$ to the sample $\mathbf{x}_{0}$ by gradually denoising $\mathbf{x}_{t}$ to less noisy samples $\mathbf{x}_{t-1}$. Formally, we are given a reverse sampling process: 
\begin{equation}
p_{\theta}\left(\mathbf{x}_{t-1} \mid \mathbf{x}_{t}\right):=\mathcal{N}\left(\mathbf{x}_{t-1} ; \boldsymbol{\mu}_{\theta}\left(\mathbf{x}_{t}, t\right), \boldsymbol{\Sigma}_{\theta}\left(\mathbf{x}_{t}, t\right)\right)
\end{equation}
\begin{equation}
\boldsymbol{\mu}_{\theta}\left(\mathbf{x}_{t}, t\right)=\frac{1}{\sqrt{\alpha_{t}}}\left(\mathbf{x}_{t}-\frac{\beta_{t}}{\sqrt{1-\bar{\alpha}_{t}}} \boldsymbol{\epsilon}_{\theta}\left(\mathbf{x}_{t}, t\right)\right)
\end{equation}

The noise predictor network $\boldsymbol{\epsilon}_{\theta}\left(\mathbf{x}_{t}, t\right)$ is used to predict the noise at step $t$, usually a parameterized variant of the U-Net architecture. The covariance predictor $\boldsymbol{\Sigma}_{\theta}\left(\mathbf{x}_{t}, t\right)$ can be set to a fixed set of scalar covariances, or it can be learned like \cite{nichol2021improved}.

\subsubsection*{Network Architecture}

In this work, we utilize the U-Net architecture, which is employed for dense predictions in medical image segmentation, as well as the noise predictor network in DDPM. Despite sharing the same U-Net architecture, we refer to it as the noise predictor U-Net and the segmentation U-Net to distinguish its roles in the upstream and the downstream task. 

We largely retain the network structure with some modifications tailored to medical imaging tasks. Specifically, we redesign the input and output layers to process single-channel CT images and predict noise for CT images instead of the typical three-channel RGB images. The basic U-Net network consists of two inputs: the noisy image $\mathbf{x}_{t}$ and the embedding vector for step $t$. For further ablation studies, we also introduce a plain U-Net variant that has the same scale of learnable parameters while removing all components related to the diffusion step. With different channel widths of the ResBlock, two network variants, U-Net (c=128) and U-Net (c=256), are trained in the following experiments.

\subsection*{Multi-organ Segmentation}

\subsubsection*{Transfer Strategy}

As shown in Figure \ref{fig1: framework} \textbf{(B)}, the transfer strategy aims to transfer the pre-trained model in DDPM to multi-organ segmentation tasks. First, we initialize the weights of segmentation U-Net from the noise predictor U-Net in DDPM. The checkpoint selection criterion assumes that optimal generation performance correlates with the quality of feature representations. Based on the results in Table \ref{table1: ddpm resolution 256}, we consequently loaded pre-trained weights from 250,000 iterations for U-Net (c=128) and 300,000 iterations for U-Net (c=256).

Second, we address the input discrepancy between the noise predictor U-Net in DDPM and the segmentation U-Net. While the noise predictor U-Net requires both the noisy image and the diffusion step $t$ as inputs, the segmentation U-Net only requires the image to be segmented. To address the difference, we treat the diffusion step value as a hyper-parameter that influences the initialization of network parameters, which determines the scale and shift for each Resblock in the network. The impact of the diffusion step value on the downstream task is significant. The experimental results in Section \ref{section: Experiments and Results} are presented to provide insights for the diffusion step setting in multi-organ segmentation tasks.

\begin{figure}[h!]
    \centering
    \includegraphics[width=0.5\textwidth]{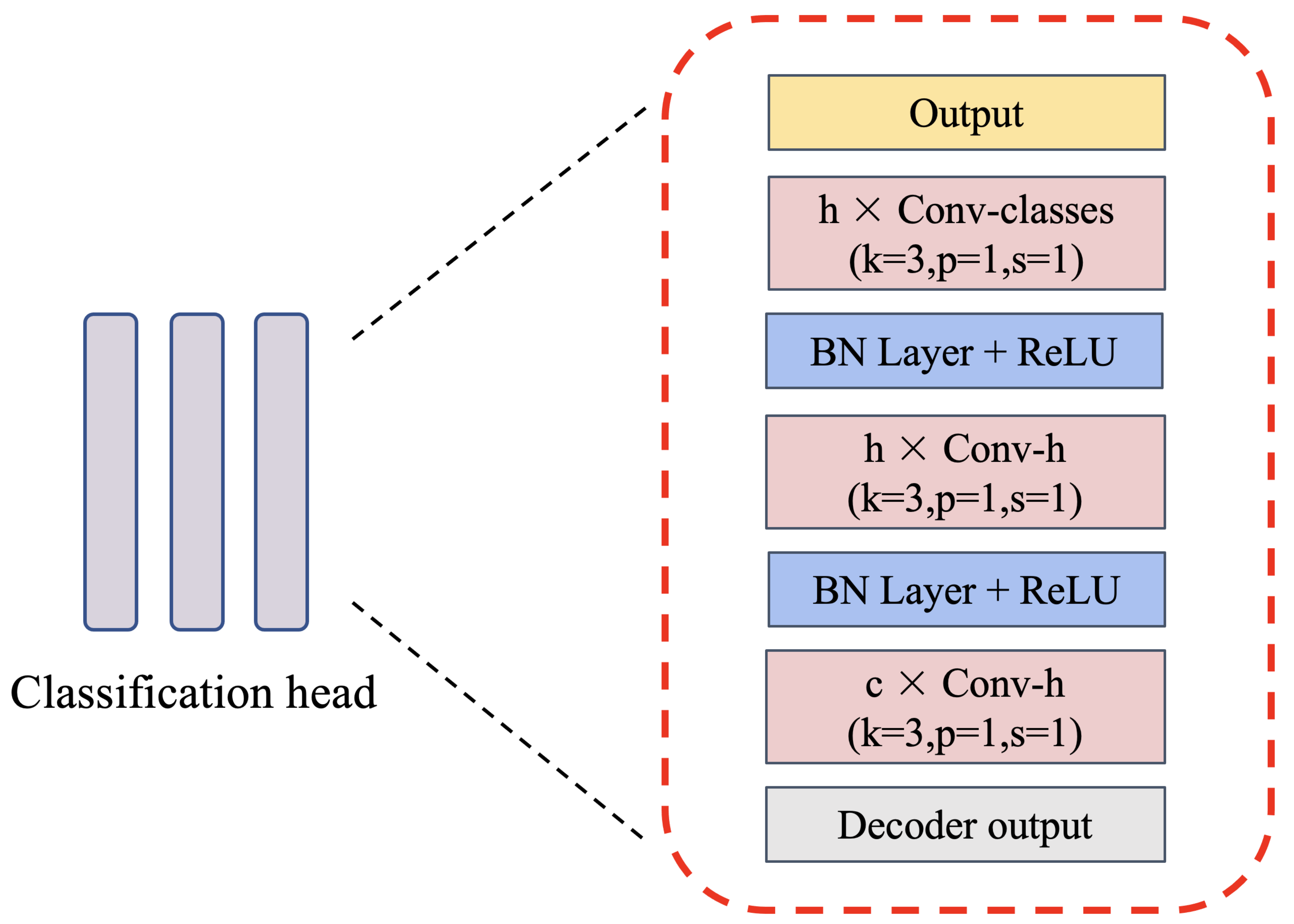}
    \caption{Classification head for segmentation tasks. c and h are the number of input channels and output channels. k, p, and s denote the convolution kernel
size, padding, and stride, respectively.}
    \label{fig2: classification_head}
\end{figure}

Third, we modify the output layer of the noise predictor U-Net to adapt it to the multi-organ segmentation task. Specifically, we redesign the classification head specifically for the target dataset and train it from scratch. The classification head (Figure \ref{fig2: classification_head}) consists of two sets of consecutive convolution layers, batch normalization (BN) layers, rectified linear unit (ReLU) activation functions, and a separate convolution layer. It involves three hyper-parameters: the number of input channels, hidden layer channels, and output channels. The number of input and output channels are determined by the ResBlock channel width and the number of organs to be segmented. The hidden layer channels are set to 128 or 256 in our experiments.

The final step is to fine-tune the pre-trained noise predictor U-Net using labeled data. Through modifications in the previous three steps, the noise predictor U-Net has been fully adapted to the segmentation tasks. Unlike traditional supervised learning methods that train the whole network from scratch, we propose lightweight fine-tuning strategies to fine-tune and update specific modules efficiently. These strategies will be detailed in the following section.

\subsubsection*{Fine-tuning strategies}

\begin{figure}[h!]
    \centering
    \includegraphics[width=\textwidth]{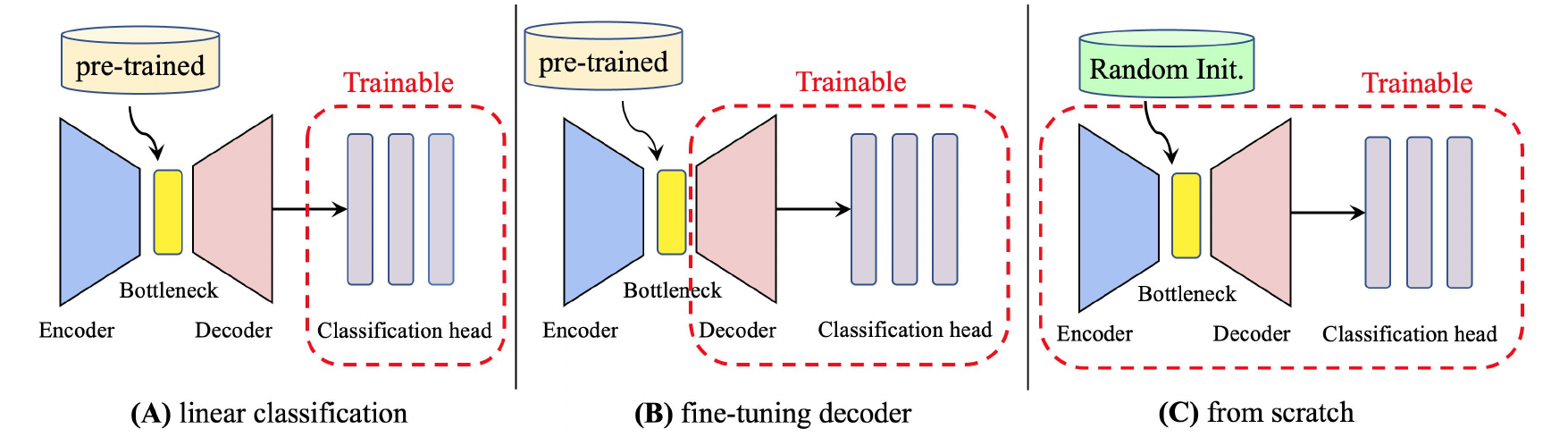}
    \caption{Fine-tuning strategies for downstream multi-organ segmentation.}
    \label{fig3: training_strategy}
\end{figure}

Figure \ref{fig3: training_strategy} illustrates three fine-tuning strategies: linear classification, fine-tuning decoder, and from scratch. Initially, we adopted a training process by using a feature extractor combined with a linear classifier, freezing the entire parameters in the backbone network, and updating only the classification head. This strategy, termed as linear classification, aimed to minimize parameter updates while leveraging pre-trained features. However, it yielded unsatisfactory results in segmentation tasks, indicating that a brute end-to-end fine-tuning is inferior to manually designed features, as demonstrated in \cite{baranchuk2021label}.

Reflecting on the performance gap between our approach and \cite{baranchuk2021label}, it is essential to guide the pre-trained model to automatically learn how to select the most important feature representations from labeled data. Thus, we explore a more effective strategy by updating additional weights, particularly targeting the decoder blocks in the U-Net, which is referred to as fine-tuning decoder. With an increase in the learnable parameters, the segmentation performance is significantly improved.

For a fair comparison with supervised approaches using the same architecture, we implemented a from scratch strategy, training models with random initialization. This strategy serves as a baseline to evaluate the effectiveness of generative pre-training and fine-tuning.

\section*{Experiments and Results}
\label{section: Experiments and Results}

\subsection*{Datasets}
We conducted our experiments using the MICCAI FLARE22 dataset, which comprises 2000 unlabeled and 50 labeled CT scans involved with 13 abdominal organs.\footnote{https://flare22.grand-challenge.org/} This dataset is the benchmark for semi-supervised abdominal multi-organ segmentation in CT images. In the experiments, we used the first 1000 unlabeled cases to pre-train the upstream noise predictor U-Net by the DDPM process, while the 50 labeled cases were split into training and set sets by a ratio of 4:1 for fine-tuning and evaluation of the downstream segmentation task.

\subsection*{Preprocessing}

We adopted a combined pipeline following \cite{isensee2021nnu}, \cite{dhariwal2021diffusion} to process the unlabeled and labeled cases separately. The preprocessing steps included (i) resampling, (ii) intensity normalization, (iii) splitting, and (iv) data augmentation.

(i) All  images and labels are resampled to a specific target spacing, using tri-linear interpolation and nearest-neighbor interpolation, respectively. The target spacing is determined based on the distribution of a specific dataset.

(ii) For intensity normalization, we compute the maximum, minimum, 0.5 percentile, and 99.5 percentile values of the voxel intensities across all unlabeled cases, including the background class rather than only foreground classes in nnU-Net. Voxel intensities of all images were then clipped to the corresponding 0.5 and 99.5 percentiles and normalized using the min-max normalization to the range [0,1]. Subsequently, the intensities are further adjusted to the range [\textendash1,1] through a linear transformation, which is consistent with the range in \cite{dhariwal2021diffusion}. The labeled images are normalized by the same pipeline, with statistics derived from the unlabeled images.

(iii) We split the CT images into 2D slices along the transverse plane, producing a total of 207,029 slices for generative tasks. For segmentation tasks, 3879 and 915 slices are used as training and test sets, respectively.

(iv) We resize images and labels (if available) to a specified resolution of 256x256 using bi-linear interpolation for images and nearest-neighbor interpolation for labels. Horizontal flipping is applied for pre-training DDPMs. No additional augmentations are used for segmentation tasks. 

\subsection*{Implementation Details}

\subsubsection*{Loss Function}

For training the noise predictor network in the DDPM, we use the mean square error (MSE) loss function to calculate the error between the real noise $y$ and the estimated noise $\hat{y}$. 

\begin{equation}
\mathcal{L}_{noise} = \frac{1}{N}\sum_{i=1}^{N}(y_i - \hat{y_i})^2
\end{equation}
We use a combination of cross entropy loss and dice loss for multi-organ segmentation tasks shown in Eq. \ref{loss}.
\begin{equation}
\label{loss}
\mathcal{L}_{seg}=w\mathcal{L}_{CE}+ (1-w)\mathcal{L}_{Dice}
\end{equation}
where $w $ is the weight of two losses set at 0.5 in our experiment.  The equations of cross entropy and dice loss are defined as:
\begin{equation}
\label{ce_loss}
\mathcal{L}_{CE}=-\frac{1}{N}\sum_{c=1}^{C}\sum_{i=1}^{N} g_{i}^{c} \log y_{i}^{c}
\end{equation}

\begin{equation}
\label{dice_loss}
\mathcal{L}_{Dice}=1-\frac{2 \sum_{c=1}^C \sum_{i=1}^N g_i^c y_i^c + \epsilon}{\sum_{c=1}^C \sum_{i=1}^N g_i^c+\sum_{c=1}^C \sum_{i=1}^N y_i^c + \epsilon}
\end{equation}
where $g_i^c$ is the ground truth binary indicator of class label $c$ of voxel $i$,  and $y_i^c$ is the corresponding predicted segmentation probability. In Eq. \ref {dice_loss}, $\epsilon$ is set to a small number to prevent the denominator from being 0, which is set as 1e-5 by default in our experiment. 

\subsubsection*{Experiment Settings}

The experiments were conducted using PyTorch framework on NVIDIA A100 GPUs. For CT image generation, we specify a resolution of 256x256, with ResBlock channel widths of 128 and 256. Gaussian noise is gradually added to the images, starting from an initial value of $\beta_1$ at 0.0001 and terminating with a $\beta_T$ value of 0.02. The sampling step, denoted as $T$, is set to 1000, with a cosine annealing scheduler employed for the scheduling strategy. We employ the exponential moving average (EMA) during the pre-training stage with a momentum value of 0.9999. The model was optimized using the Adam optimizer for 300,000 training iterations, with a linearly decreasing learning rate from the initial value of 2e-4 to 2e-5.

As for the fine-tuning settings for segmentation, we maintained consistency by employing the same optimizer across fine-tuning strategies. Through extensive experiments, we found that the network achieves optimal performance when the learning rate of the classification head is ten times higher than the base rate. Additionally, we found that appropriate weight decay regularization contributes to both model convergence and performance improvement. Consequently, we used the Adam optimizer and set weight decay value to 1e-3 for the classification head and 1e-4 for the remaining components in network. The model is fine-tuned for a total of 30,000 iterations in the fully-data setting, while 10,000 iterations in the label-efficient learning setting.

\subsubsection*{Evaluation Metrics}

To assess the quality of CT images generated by DDPM, we use several metrics for generative models, including Fréchet inception distance (FID) \cite{heusel2017gans}, spatial Fréchet inception distance (sFID) \cite{nash2021generating}, precision, recall, and F1-score \cite{kynkaanniemi2019improved}. Specifically, we selected 2000 real CT images from the unlabeled cases in the FLARE22 dataset to compare its distribution with that of the generated samples using the aforementioned metrics. We use the dice similarity coefficient (DSC) and the Jaccard index (JI) to evaluate the performance of methods on medical image segmentation tasks. 

\subsection*{CT Image Synthesis Performance}

The quality of CT images generated by DDPM was evaluated on two different scales, U-Net (c=128) and U-Net (c=256). For each DDPM model, 300,000 iterations are trained and evaluated on FID, sFID, precision, recall, and F1-score metrics every 50,000 iterations.

\begin{figure}[h!]
    \centering
    \includegraphics[width=\textwidth]{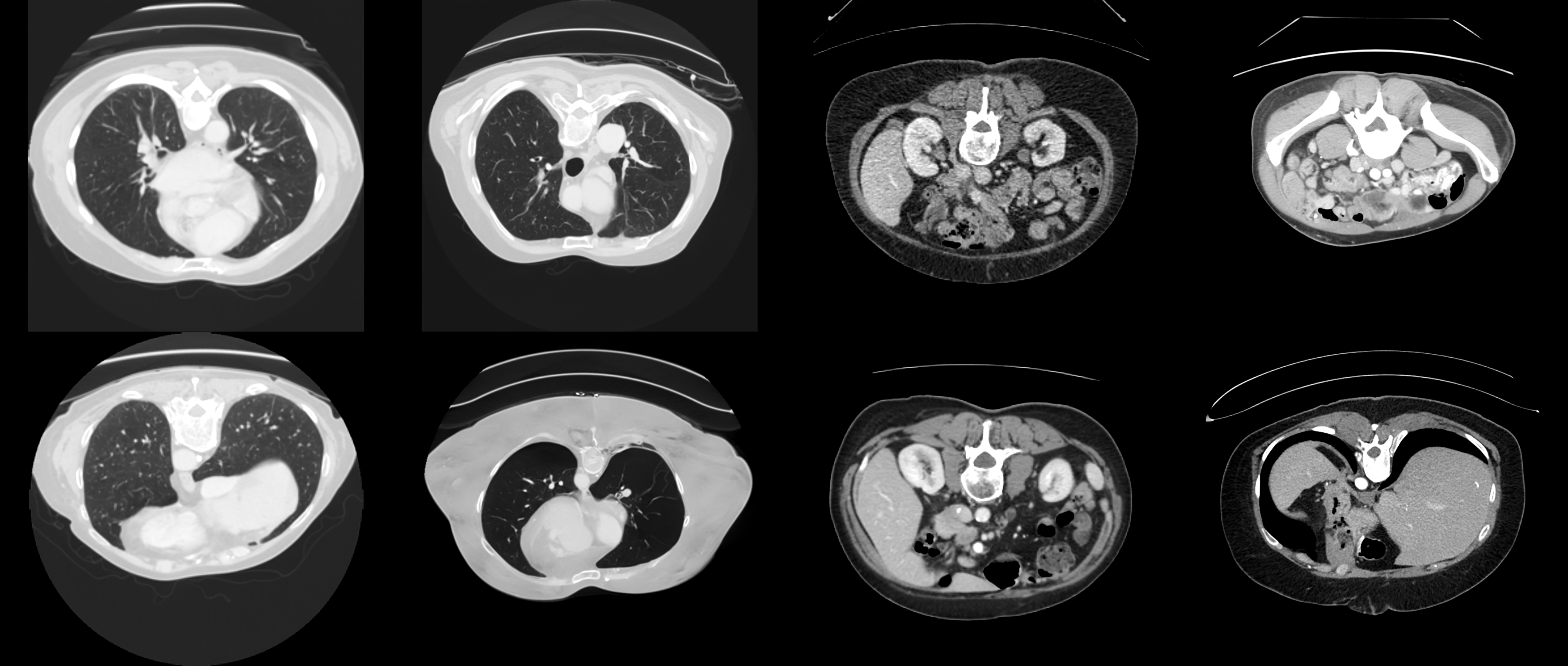}
    \caption{Samples generated by DDPM in lung view(W:1400, L:-500, first two columns) and abdominal view (W:350, L:40, last two columns).}
    \label{fig4: generated_samples}
\end{figure}

Table \ref{table1: ddpm resolution 256} demonstrates that the variant U-Net (c=128) outperforms the U-Net (c=256) in generative tasks. This suggests that the parameter scale is not the primary factor in optimizing the generative model in this experiment. Although the larger network converges more slowly and exhibits slightly lower performance than the smaller network, both models can generate diverse and high-quality 2D CT images. In Figure \ref{fig4: generated_samples}, some representative samples generated by the optimal checkpoint of U-Net (c=128, iteration=250,000) are shown in lung and abdominal view, including most of the categories of abdominal organs.

For the generation task at a resolution of 256x256, the smaller U-Net network tends to stabilize after 150,000 training iterations. The model with 250,000 iterations performs the best, achieving the highest level for 4 out of the 5 metrics. However, it is worth noting that the precision reaches its optimal value after 300,000 iterations. It can also be observed that training a larger network U-Net (c=256) poses more challenges and exhibits slower convergence. These larger-scale networks stabilize after approximately 200,000 training iterations, with the optimal model achieved at 300,000 iterations. Although precision falls slightly below optimal level, the remaining four indicators perform optimally.

\subsection*{Ablation Experiments for Multi-organ Segmentation}

To assess the segmentation performance of our proposed method on the FLARE22 dataset, we investigate the impact of model scales, fine-tuning strategies, and initialization diffusion steps. For the fine-tuning decoder fine-tuning strategy and linear classification, we employ 11 initialization diffusion steps. These steps involve selecting values for $t$ ranging from 0 to 1000 at intervals of 100.

To investigate the impact of different ResBlock widths in the backbone and the number of channels in the classification head, three parallel experiments were conducted. These experiments used the same network architecture but varied in scale. These experiments are denoted as model Small (S), Medium (M) and Large (L). The channel widths and hidden layers for the three models are set as follows: 128 + 128 for Small, 128 + 256 for Medium, and 256 + 256 for Large. 


Based on the experimental results presented in Table \ref{table2: Quantitative Results on FLARE22 Dataset}, it is observed that the plain U-Net network trained through supervised learning achieved an average DSC score of approximately 80\% and an average JI score above 70\%. This indicates that the U-Net network still performs well in segmentation tasks even after removing the structure related to the diffusion step. This demonstrates the effective transferability of networks between noise prediction and segmentation tasks. The enhancements made to improve the performance of generative models also provide benefits in terms of segmentation performance. Moreover, based on this result, it can be concluded that the number of channels is not the limiting factor for further improvement in segmentation performance. This is evidence that model M, with more channels than model S, performs worse in segmentation performance. 


For the other two proposed fine-tuning strategies, it was observed that under the linear classification strategies, the segmentation performance was extremely poor. In fact, training even failed for most of the initialization diffusion steps, resulting in an average DSC and JI of less than 28.85\% and 22.34\%, respectively. These results demonstrate that the features learned from the proposed upstream pre-training task cannot be directly used as semantic features in the downstream segmentation task. Simply updating the parameters in the classification head hinders the pre-trained models from effectively adapting to the downstream tasks, leading to poor performance due to a lack of expressive power in the features. In other words, the knowledge learned from the pre-trained model requires a more extensive fine-tuning process rather than solely relying on the linear probing method employed in other pre-training tasks. This fundamental reason motivated us to implement the fine-tuning decoder strategy. Additionally, it is worth noting that despite this result not being applicable to segmentation, we arrive at a similar conclusion as in \cite{baranchuk2021label}: the optimal diffusion step for obtaining semantic features generally falls within the range of 0 to 400. 


Through unfreezing the decoder in the pre-trained model, a notable improvement in the overall performance is observed. Specifically, in model S, the fine-tuning decoder strategy outperformed the supervised learning method with the same network architecture by 6.32\% and 5.54\% in terms of DSC and JI, respectively. However, as the network scale increased, the improvement achieved by fine-tuning decoder-based networks gradually decreased. Model L showed no difference between the two fine-tuning strategies, indicating that transferring large-scale networks is more challenging than smaller ones. Across all experimental settings, the best and sub-optimal segmentation models were model S and model L, respectively, initialized with a diffusion step of 0. These models achieved DSC scores of up to 86.91\% and 85.21\%, and JI scores of 80.38\% and 78.66\%, respectively. By examining the relationship between the segmentation performance and the initialization diffusion step, we can infer that the optimal initialization step generally falls within the range of 0 and 300. 

\subsection*{Multi-organ Segmentation Performance on FLARE Dataset}

\subsubsection*{Comparison with Existing Methods}


We evaluated several supervised learning approaches on the FLARE22 dataset to compare our method with other advanced multi-organ segmentation methods. These architectures include DeepLabV3+ \cite{chen2018encoder}, U-Net \cite{ronneberger2015u} and its variants with different backbones, ResU-Net \cite{diakogiannis2020resunet}, U-Net++ \cite{zhou2019unet++}, Attention U-Net \cite{oktay2018attention}, as well as transformer-based architectures such as UNETR \cite{hatamizadeh2022unetr} and Swin UNETR \cite{hatamizadeh2022swin}. Additionally, we compared our method with nnU-Net \cite{isensee2021nnu}, a representative medical image segmentation framework, and DDPM-Seg \cite{baranchuk2021label}, another diffusion model-based segmentation method.

For our evaluation, we used the public implementations of the DeepLabV3+, ResU-Net, and U-Net++ pre-trained on ImageNet. These implementations were obtained from this repository.\footnote{https://github.com/qubvel/segmentation\_models.pytorch} We also used the implementations of Attention U-Net, UNETR, and Swin UNETR from the MONAI project.\footnote{https://github.com/Project-MONAI/MONAI}

\subsubsection*{Analysis}

The results of the quantitative evaluation on the FLARE22 dataset are presented in Table \ref{table3: Comparison to Other SOTA Methods}. Comparing the first three methods, DeepLabV3+, ResU-Net, and U-Net++ with ImageNet pre-training weights, we can infer that the larger-scale U-Net++ performs worse than ResU-Net. This suggests that the features learned from ImageNet pre-training are unsuited for CT images, indicating a domain gap between general image data and medical images. Furthermore, we evaluated three network architectures that were randomly initialized: Attention U-Net, UNETR, and Swin UNETR. Like the ImageNet pre-trained methods, the best-performing method among these three, Attention U-Net, achieved a DSC score of less than 80\%. These results demonstrate that these methods are insufficient for multi-organ segmentation without extensive data augmentation and carefully designed training techniques specific to the downstream task.


The results in the last seven rows show the performance of the proposed method with three different fine-tuning strategies. The linear classification fine-tuning strategy exhibited poor performance. However, the fine-tuning decoder strategy achieved DSC and JI scores of 86.91\% and 80.38\% in model S, surpassing the supervised learning method with the same network architecture. These results indicate that the proposed method with the fine-tuning decoder strategy significantly outperforms existing supervised segmentation methods, except for 2D nnU-Net. However, through our proposed approach performs slightly inferior to 2D nnU-Net when labeled data are abundant, the performance of 2D nnU-Net drops dramatically when labeled data is scarce, whereas our method remains superior under the same conditions. 

\subsection*{Label-efficient Learning}

\subsubsection*{Competing Methods}

To further evaluate the performance of the proposed method under conditions of limited data availability, we conducted experiments using three different levels of labeled data: 1\% (39 slices), 10\% (388 slices), and only one batch (approximately 0.1\%, 4 slices). For the experiments with 1\% and 10\% data, we randomly selected sub-datasets without any manual screening to ensure fairness. However, for the experiments using only one batch, it was necessary to carefully check the sub-dataset to ensure that each organ category appeared at least once. Failure to include certain organs in this small sub-dataset could lead to unsuccessful segmentation by the methods. In these experiments, we followed the same settings as the full dataset, except for limiting the fine-tuning decoder strategy to 1000 iterations when using only one batch of data. This adjustment aims to prevent overfitting with limited training data. 


To reproduce the work of DDPM-Seg on the FLARE22 dataset, we use the same DDPM pre-trained weight as ours and follow the guidance carefully by default settings in DDPM-Seg. Specifically, the dimension of pixel-level representations is 8448, which are from the middle blocks of the UNet (c=256) decoder B= \{5, 6, 7, 8, 12\} and diffusion steps t = \{50, 150, 250\}. Additionally, we halve the dimension of representations to utilize the pre-trained U-Net (c=128). We also train an ensemble of independent MLPs using these features. In addition, DDPM-Seg is a RAM-consuming method for label-efficient segmentation since it keeps all training pixel representations in memory, which requires more than 210 GB for 50 training images of resolution 256x256.\footnote{https://github.com/yandex-research/ddpm-segmentation} Therefore, we perform DDPM-Seg only under conditions of 0.1\% and 1\% of labeled data.

To implement nnU-Net under the conditions of 1\% and 10\% of labeled data, we subset the data in units of cases from the FLARE2022 dataset, instead of splitting in units of slices, which is in line with nnU-Net's pipeline. Specifically, we consider a single case and five cases as an independent dataset, using nnU-Net to train and infer segmentation masks. Due to nnU-Net requiring at least one complete 3D CT image, we cannot perform nnU-Net under 0.1\% of labeled data. As a result, some results are marked as "NA" in Table \ref{table3: Comparison to Other SOTA Methods} when nnU-Net and DDPM-Seg do not support the conditions discussed above.

\subsubsection*{Analysis}

As shown in Table \ref{table4: label-efficient}, the fine-tuning decoder strategy consistently outperforms the nnU-Net and DDPM-Seg methods across all three ratios of labeled data. Notably, our method exhibits a substantial performance advantage, particularly when working with limited labeled data. Compared to other segmentation methods, the gap in performance between our method and the alternatives becomes more significant as the ratio of labeled data decreases. The performance degradation is not markedly different under the 10\% labeled data condition. However, our method consistently outperforms the others by a larger margin when the labeled dataset is reduced to only 1\% or even as low as 0.1\%. These conclusions are supported by both the DSC and the JI metrics, as indicated in Table \ref{table4: label-efficient}. In the following, we primarily focus on the DSC metric to highlight the strengths of our approach over the other approaches.

In-depth analysis of the results reveals that the proposed method performs consistently well, using large-scale or small-scale datasets with only a few labeled samples. Under the 0.1\%, 1\%, and 10\% labeled data conditions, the proposed method achieved DSCs of 51.81\%, 71.56\%, and 78.51\%, and JIs of 64.21\%, 72.43\%, and 78.51\%, respectively. Compared to the SOTA methods, nnU-Net and DDPM-Seg, the proposed method significantly improves segmentation performance on small-scale datasets. Specifically, the fine-tuning strategy outperforms 2D nnU-Net by 12.87\% and 5.08\% in terms of DSC under the 1\% and 10\% labeled data conditions. Furthermore, it surpasses DDPM-Seg by 8.42\% and 10.78\% under the 0.1\% and 1\% labeled data conditions, respectively. It is worth noting that, compared to using all available data, the proposed method experiences a minor reduction in DSC  (35.1\% and 15.35\%) under the 0.1\% and 1\% labeled data conditions. In contrast, the other methods exhibit a larger decline, ranging from 45.65\% to 56.22\% and 24.29\% to 35.93\%. These results highlight the competitive performance of the proposed method and its ability to address the limitations of nnU-Net and DDPM-Seg, extending its applicability to a wider range of applications. 

Comparing the performance of the fine-tuning strategy (rows 9\textendash11) with competing methods (rows 1\textendash8), it is notable that the improvement achieved in the full data settings diminishes significantly under the 0.1\% labeled data condition. However, the fine-tuning decoder strategy remains competitive under the same condition as before, indicating that the improvement offered by our method is primarily derived from the generative pre-training by DDPM and the fine-tuning decoder strategy rather than any architectural advances.

\begin{figure}[h!]
    \centering
    \includegraphics[width=\textwidth]{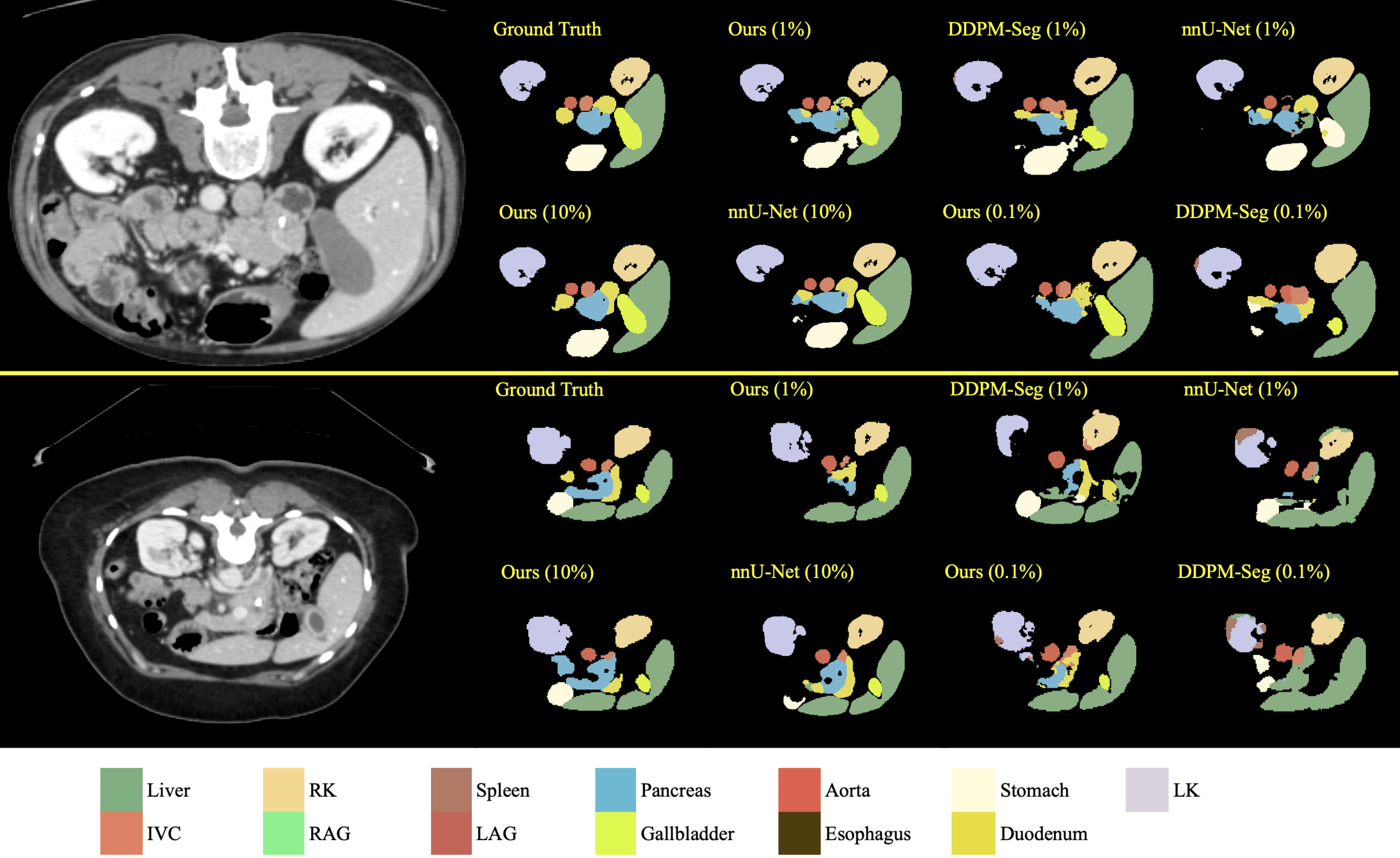}
    \caption{Visualization of segmentation performance with different fine-tuning strategies across labeled data ratios.}
    \label{fig5: visualization_seg}
\end{figure}

Furthermore, we explored the segmentation performance of these approaches in all organs separately. Organ-level results are provided in Table \ref{table5: label-efficient organ-level results}. Intuitively, it can be inferred that our proposed method (fine-tuning decoder) achieves the best segmentation performance for most organs under three data ratio settings. Compared to these SOTA methods, organ-level results also reveal that the proposed method has good robustness of segmentation performance for abdominal organs. However, some cases show that the segmentation improvement of the proposed method for smaller organs is not as stable as that for large ones. For example, our proposed method, fine-tuning decoder S, has a sharp decrease compared to nnU-Net under 10\% labeled data. Additionally, the visualization results of the multi-organ segmentation on the FLARE22 test set are given in Figure \ref{fig5: visualization_seg}.

\section*{Discussion}
Although the experimental results highlight the advantages of our proposed method in label-efficient learning, some aspects require further exploration and refinement.

First, the linear classification strategy, commonly employed in self-supervised methods for downstream tasks, showed limited adaptability to multi-organ segmentation in our experiments. This challenge may arise from the distribution of semantic features within the deep layers of the network. Our analysis of DDPM-Seg indicates that the most informative features tend to emerge in the middle layers of the U-Net decoder. Since our current approach relies only on the final feature map, it may not fully leverage the richness of intermediate feature representations. A potential improvement is to integrate feature maps from multiple layers within the U-Net, enabling a more comprehensive representation for segmentation.

Second, the current implementation focuses on 2D models at a resolution of 256×256. While this resolution effectively captures essential structures, high-resolution synthesis using DDPM remains a challenging task, particularly for 3D medical imaging modalities like CT and MRI. Training high-quality generative models at resolutions of 512×512 or beyond requires additional optimization and computing resources. Future work could explore advanced diffusion models tailored for high-resolution and volumetric medical imaging, enabling improved generalization across different imaging modalities.

Lastly, the role of the diffusion step in downstream segmentation tasks warrants a more in-depth investigation. Our findings suggest that the choice of diffusion step significantly influences segmentation performance. However, unlike its well-defined role in controlling noise levels during DDPM training, its interpretation in downstream applications remains less clear. Establishing a theoretical framework to understand the impact of diffusion steps in specific tasks could enhance the development of more efficient and adaptable pre-trained models. Future research can draw parallels with generative models such as GANs to explore diffusion mechanisms that optimize task-specific feature representations.

\section*{Conclusions}

In this work, we propose a novel approach for multi-organ segmentation by leveraging a pre-trained DDPM-based model. Through pre-training a U-Net architecture following the DDPM process, the model not only generates high-quality synthetic CT images but also captures robust and meaningful representations from unlabeled CT data, which can be effectively transferred to downstream segmentation tasks. By introducing lightweight transfer and fine-tuning strategies, we bridge the gap between generative pre-training by DDPM and medical image segmentation. Extensive experiments on the FLARE22 benchmark demonstrate that our approach outperforms state-of-the-art supervised approaches, particularly in low-data scenarios. By minimizing the reliance on large-scale labeled datasets, our approach provides a practical and label-efficient solution for CT organ segmentation, highlighting the potential of diffusion-based pretraining for advancing medical image analysis.




\section*{Author Contributions}
\textbf{YH:} Conceptualization, Formal analysis, Investigation, Theory \& methodology, Project administration, Resources, Software, Validation, Writing – review \& editing. \textbf{FX:} Data curation, Formal analysis, Investigation, Validation, Visualization, Clinical validation, Review \& editing. \textbf{LT:} Methodology, Investigation, Results discussion, Funding acquisition, Review \& editing, Supervision. \textbf{JZ:} Results discussion, Review \& editing. \textbf{HH:} Investigation, Clinical application, Review \& editing. \textbf{LS:} Methodology, Verified the theory and analytical results, Review \& editing. \textbf{YP:} Data curation, Clinical validation, Investigation, Results discussion, Review \& editing. \textbf{JL:} Writing – review \& editing. \textbf{JM:} Data curation, Clinical validation, Results discussion, Supervision, Review \& editing. \textbf{BH:} Conceptualization, Formal analysis, Results discussion, Project administration, Supervision, Validation, Writing – review \& editing.


\bibliography{ref}

\section*{Figures and Tables}

\begin{table}[h!]
\begin{minipage}{0.5\textwidth}
\caption{Evaluation on generated CT images of a resolution of 256x256.}
\label{table1: ddpm resolution 256}
\resizebox{\textwidth}{!}{%
\begin{tabular}{llcccccc}
\toprule
                       &    & 50,000     & 100,000    & 15,000    & 200,000    & 250,000    & 300,000    \\ \midrule
\multirow{5}{*}{c=128} & FID       & 31.4092 & 16.3188 & 12.0206 & 11.5218 & \textbf{11.3162} & 12.0449 \\
                       & sFID      & 72.5776 & 56.0745 & 48.8117 & 47.5353 & \textbf{46.9282} & 47.4179 \\
                       & Precision & 0.489   & 0.6755  & 0.758   & 0.7995  & 0.796   & \textbf{0.803}   \\
                       & Recall    & 0.4035  & 0.5635  & 0.585   & 0.63    & \textbf{0.676}   & 0.659   \\
                       & F1-score  & 0.4422  & 0.6144  & 0.6604  & 0.7047  & \textbf{0. 7311} & 0.7239  \\ \midrule
\multirow{5}{*}{c=256} & FID       & 38.8874 & 38.5144 & 24.4172 & 13.4527 & 13.479  & \textbf{12.2408} \\
                       & sFID      & 83.7842 & 85.2128 & 63.9002 & 52.5795 & 53.1192 & \textbf{50.3049} \\
                       & Precision & 0.3765  & 0.434   & 0.5985  & 0.7135  & \textbf{0.736}   & 0.724   \\
                       & Recall    & 0.319   & 0.535   & 0.588   & 0.599   & 0.595   & \textbf{0.642}   \\
                       & F1-score  & 0.3454  & 0.4792  & 0.5932  & 0.6513  & 0.658   & \textbf{0.6805}  \\ \bottomrule
\end{tabular}%
}
\end{minipage}
\end{table}

\begin{table}[h!]
\begin{minipage}{0.5\textwidth}
\centering
\caption{Quantitative results on the FLARE22 dataset with different fine-tuning strategies.}
\label{table2: Quantitative Results on FLARE22 Dataset}
\resizebox{\textwidth}{!}{%
\begin{tabular}{@{}lccccccc@{}}
\toprule
\multirow{2}{*}{Strategies} & \multirow{2}{*}{Step} & \multicolumn{3}{c}{Dice Coef.(\%)} & \multicolumn{3}{c}{Jaccard Index (\%)} \\ \cmidrule(l){3-8}
                                       &        & S     & M     &  L     & S     & M     &  L     \\ \midrule
Scratch.                           & -      & 80.59 & 79.41 & 83.83 & 74.72 & 73.78 & \textbf{77.16} \\ \midrule
\multirow{5}{*}{Linear.} & 100    & 26.86 & 26.29 & 10.63 & 20.24 & 19.86 & 7.81  \\
                                       & 200    & 28.85 & 20.41 & 13.84 & 22.19 & 13.9  & 10.78 \\
                                       & 300    & 18.72 & 22.45 & 17.98 & 13.76 & 16.4  & 13.61 \\
                                       & 400    & 28.23 & 23.72 & 17.98 & 22.34 & 17.93 & 13.61 \\ \cmidrule(l){2-8}
                                       & Others & \multicolumn{6}{c}{Failed}                    \\ \midrule
\multirow{11}{*}{Fine-tuning.}  & 0      & \textbf{86.91} & \textbf{85.21} & 79.76 & \textbf{80.38} & \textbf{78.66} & 74.17 \\
                                       & 100    & 77.98 & 79.29 & 78.32 & 72.13 & 73.31 & 72.43 \\
                                       & 200    & 77.15 & 77.29 & 78.71 & 71.22 & 71.21 & 72.71 \\
                                       & 300    & 83.73 & 76.5  & \textbf{83.84 }& 76.92 & 70.55 & 76.87 \\
                                       & 400    & 76.94 & 81.28 & 78.86 & 70.54 & 74.04 & 72.8  \\
                                       & 500    & 81.54 & 76.71 & 77.01 & 74.42 & 70.33 & 70.68 \\
                                       & 600    & 80.25 & 73.87 & 77.8  & 72.65 & 67.3  & 71.78 \\
                                       & 700    & 72.48 & 73.03 & 76.3  & 65.69 & 66.08 & 69.83 \\
                                       & 800    & 68.95 & 68.98 & 74.44 & 61.79 & 61.9  & 67.86 \\
                                       & 900    & 63.94 & 64.18 & 74.71 & 57.86 & 57.73 & 67.92 \\
                                       & 1000   & 64.47 & 69.31 & 74.64 & 58.07 & 62.48 & 67.65 \\ \bottomrule
\end{tabular}
}
\end{minipage}
\end{table}

\begin{table}[h!]
\begin{minipage}{0.5\textwidth}
\centering
\caption{Performance comparison between existing methods for CT multi-organ segmentation on the FLARE22 dataset. *: The optimal diffusion step is 300 (default 0 if required).}
\label{table3: Comparison to Other SOTA Methods}
\resizebox{\columnwidth}{!}{%
\begin{tabular}{@{}lcccc@{}}
\toprule
Method                             & W/ unlabeled & Pre-trained    & DSC (\%) & JI (\%) \\ \midrule
DeepLabV3+                         & No           & ImageNet & 67.75           & 58.19              \\
ResU-Net                           & No           & ImageNet & 77.22           & 69.3               \\
U-Net++                            & No           & ImageNet & 65.28           & 57.31              \\ \midrule
Attention U-Net                    & No           & \textemdash            & 77.07           & 68.64              \\
UNETR                              & No           & \textemdash             & 64.72           & 54.62              \\
Swin UNETR                          & No           & \textemdash             & 73.86           & 64.83              \\ \midrule
nnU-Net 2D                         & No           & \textemdash             & 87.39           & 81.72              \\ \midrule
Linear. (ours) & Yes          & DDPM  & 28.85           & 22.19              \\ \midrule
Scratch. S (ours)          & No           & \textemdash             & 80.59           & 74.72              \\
Scratch. M (ours)          & No           & \textemdash             & 79.41           & 73.78              \\
Scratch. L (ours)          & No           & \textemdash             & 83.83           & 77.16              \\ \midrule
Fine-tuning. S   (ours) & Yes          & DDPM  & 86.91           & 80.38              \\
Fine-tuning. M   (ours) & Yes          & DDPM  & 85.21           & 78.66              \\
Fine-tuning. L   (ours) & Yes          & DDPM  & 83.84*        & 76.87              \\ \bottomrule
\end{tabular}%
}
\end{minipage}
\end{table}

\begin{table*}[h!]
\begin{minipage}{\textwidth}
\centering
\caption{Performance comparison between different methods for CT multi-organ segmentation under different labeled data ratios. Values in the bracket indicate the gap compared with the performance of corresponding models under full data. *: The optimal diffusion step is 200 (default 0 if required).}
\label{table4: label-efficient}
\resizebox{\textwidth}{!}{%
\begin{tabular}{lcccccc}
\toprule
                         & \multicolumn{3}{c}{Dice coef. (\%)}              & \multicolumn{3}{c}{Jaccard Index (\%)}           \\ \cline{2-7} 
\multirow{-2}{*}{Method} & $\sim$0.1\%    & 1\%            & 10\%           & $\sim$0.1\%    & 1\%            & 10\%           \\ \hline
DeepLabV3+                   & 20.74 (\textendash47.01) & 41.78 (\textendash25.97) & 58.71 (\textendash9.04)   & 15.13 (\textendash43.06) & 34.84 (\textendash23.35) & 50.52 (\textendash7.67)   \\
ResU-Net                 & 21.00 (\textendash56.22) & 41.29 (\textendash35.93) & 71.62 (\textendash5.6)   & 16.06 (\textendash53.24) & 35.99 (\textendash33.31) & 63.47 (\textendash5.83)  \\
U-Net++                  & 15.13 (\textendash50.15) & 34.22 (\textendash31.06) & 64.99 (\textendash0.29)  & 11.45 (\textendash45.86) & 29.28 (\textendash28.03) & 56.58 (\textendash0.73)  \\
Attention U-Net          & 28.81 (\textendash48.26) & 50.7 (\textendash26.37)  & 71.78 (\textendash5.29)  & 21.93 (\textendash46.71) & 42.74 (\textendash25.9)  & 63.16 (\textendash5.48)  \\
UNETR                    & 13.87 (\textendash50.85) & 33.41 (\textendash31.31) & 54.91 (\textendash9.81)  & 9.55 (\textendash45.07)  & 26.31 (\textendash28.31) & 45.15 (\textendash9.47)  \\
Swin UNETR               & 28.21 (\textendash45.65) & 49.57 (\textendash24.29) & 70.19 (\textendash3.67)  & 21.88 (\textendash42.95) & 42.06 (\textendash22.77) & 61.43 (\textendash3.40)  \\ \hline
nnU-Net               & NA             & 58.69 (\textendash28.41) & 73.43 (\textendash13.67) & NA             & 52.03 (\textendash29.69) & 66.75 (\textendash14.97) \\
DDPM-Seg (c=128)                 & 44.54          & 59.27          & NA             & 36.59          & 51.13          & NA             \\ 
DDPM-Seg (c=256)                & 43.39          & 60.78          & NA             & 35.73          & 52.65          & NA             \\ \hline
From-scratch S (ours)    & 28.34 (\textendash52.25) & 60.07 (\textendash20.52) & 69.92 (\textendash10.67) & 23.27 (\textendash51.45) & 52.24 (\textendash22.48) & 64.26 (\textendash10.46) \\
From-scratch M (ours)    & 24.68 (\textendash54.73) & 58.1 (\textendash21.31)  & 68.23 (\textendash11.18) & 19.82 (\textendash53.96) & 50.94 (\textendash22.84) & 62.51 (\textendash11.27) \\
From-scratch L (ours)    & 28.92 (\textendash54.91) & 54.71 (\textendash29.12) & 70.46 (\textendash13.37) & 24.01 (\textendash53.15) & 47.64 (\textendash29.52) & 64.97 (\textendash12.19) \\ \hline
Fine-tuning decoder S (ours) & \textbf{51.81} (\textendash35.10)                        & \textbf{71.56} (\textendash15.35) & \textbf{78.51} (\textendash8.4)    & \textbf{44.79} (\textendash35.59) & \textbf{64.21} (\textendash16.17) & \textbf{72.43} (\textendash7.95)   \\
Fine-tuning decoder M (ours) & 51.17 (\textendash34.04)                        & 70.25 (\textendash14.96) & 76.52 (\textendash8.69)* & 44.61 (\textendash34.05) & 63.31 (\textendash15.35) & 69.86 (\textendash8.80)* \\
Fine-tuning decoder L (ours) & 50.35 (\textendash33.49)                        & 69.07 (\textendash14.77) & 77.33 (\textendash6.51)   & 43.22 (\textendash33.65) & 61.93 (\textendash14.94) & 71.23 (\textendash5.64)   \\ \bottomrule
\end{tabular}%
}
\end{minipage}
\end{table*}

\begin{table*}[h!]
\begin{minipage}{\textwidth}
\centering
\caption{Organ-level DSC score between different methods for CT multi-organ segmentation under different labeled data ratios. Abbreviations: RK - Right Kidney, IVC - Inferior Vena Cava, RAG - Right Adrenal Gland, LAG - Left Adrenal Gland, and LK - Left Kidney. 'NA' denotes the DSC score using the corresponding approach for specific organs is below 1\%.}
\label{table5: label-efficient organ-level results}
\resizebox{\textwidth}{!}{%
\begin{tabular}{@{}clccccccccccccc@{}}
\toprule
Ratio                        & \multicolumn{1}{c}{Methods}  & Liver          & RK             & Spleen         & Pancreas       & Aorta          & IVC            & RAG            & LAG            & Gallbladder    & Esophagus      & Stomach        & Duodenum       & LK             \\ \midrule
\multirow{3}{*}{10\%}        & nnU-Net                      & 90.02          & 87.92          & 91.77          & 43.63          & 94.05          & 82.78          & \textbf{62.68} & 60.2           & 51.37          & 76.09          & 69.93          & 55.33          & 88.8           \\
                             & Scratch. S (ours)        & 92.28          & 94.52          & 93.31          & 60.65          & 94.42          & 87.06          & NA             & NA             & 71.17          & 82.03          & 77.65          & 62.03          & 93.89          \\
                             & Fine-tuning. S (ours) & \textbf{95.05} & \textbf{95.5}  & \textbf{94.73} & \textbf{73.95} & \textbf{94.53} & \textbf{89.03} & NA             & \textbf{72.29} & \textbf{76.86} & \textbf{83.57} & \textbf{85.88} & \textbf{64.92} & \textbf{94.31} \\ \midrule
\multirow{5}{*}{1\%}         & DDPM-Seg (c=128)             & 92.11          & 87.14          & 87.97          & 34.27          & 83.24          & 70.77          & 19.58          & 6.66           & 46.82          & 55.01          & 71.31          & 26.41          & 89.17          \\
                             & DDPM-Seg   (c=256)           & 92.76          & 85.96          & 88.9           & 35.97          & 87.8           & 72.34          & 10.45          & 23.42          & 38.82          & 57.83          & \textbf{75.15}          & 30.0             & 90.76          \\
                             & nnU-Net                      & 82.91          & 84.01          & 82.23          & 31.02          & 82.56          & 71.34          & 38.03          & 33.09          & 13.06          & 61.49          & 65.18          & \textbf{35.63}          & 82.4           \\
                             & Scratch. S (ours)        & 88.78          & 85.17          & 83.76          & 30.56          & 87.69          & 71.94          & 37.7           & 39.33          & 53.6           & 43.31          & 53.18          & 18.3           & 87.55          \\
                             & Fine-tuning. S (ours) & \textbf{94.14}          & \textbf{93.69}          & \textbf{89.33}          & \textbf{41.67}          & \textbf{93.4}           & \textbf{83.2}           & \textbf{55.56}          & \textbf{53.23}          & \textbf{66.77}          & \textbf{63.68}          & 72.18          & 30.18          & \textbf{93.14}          \\ \midrule
\multirow{4}{*}{$\sim$0.1\%} & DDPM-Seg (c=128)             & 85.85          & 78.21          & 75.37          & 22.95          & 76.6           & 59.51          & NA             & NA             & 20.46          & 5.96           & \textbf{44.98} & \textbf{36.98} & 72.21          \\
                             & DDPM-Seg   (c=256)           & 85.31          & 78.52          & 78.16          & 26.65          & 75.47          & 55.01          & NA             & NA             & 21.37          & 9.7            & 27.65          & 31.49          & 74.79          \\
                             & Scratch. S (ours)        & 66.61          & 54.51          & 50.26          & 5.79           & 55.04          & 35.46          & NA             & NA             & 21.54          & NA             & 4.71           & 9.61           & 64.11          \\
                             & Fine-tuning. S (ours) & \textbf{90.29} & \textbf{89.34} & \textbf{79.46} & \textbf{37.27} & \textbf{86.32} & \textbf{62.25} & NA             & \textbf{3.61}  & \textbf{56.26} & \textbf{13.95} & 38.02          & 28.49          & \textbf{87.23} \\ \bottomrule
\end{tabular}%
}
\end{minipage}
\end{table*}

\end{document}